\documentclass[runningheads]{llncs}

\usepackage{url}
\usepackage{amssymb}
\usepackage{amsmath}
\usepackage{bm}
\usepackage{hyperref}

\usepackage{hypcap}
\usepackage{lipsum}
\usepackage{cite}
\usepackage[linesnumbered,ruled,vlined]{algorithm2e} 
\usepackage{caption}
\usepackage{subfig}
\usepackage{mathtools}
\usepackage{algorithmic}
\usepackage{graphicx}
\begin{document}
\title{FedNet2Net: Saving Communication and Computations in Federated Learning with Model Growing\thanks{Partially supported by a DOD contract to the University of Maryland Institute for Advanced Computer Studies.}}

\vspace{-3mm}
\author{Amit Kumar Kundu 
\and
Joseph Jaja 
}
\authorrunning{A. K. Kundu, and J. JaJa}
%
\institute{University of Maryland, College Park, MD 20742, USA\\
\email{\{amit314,josephj\}@umd.edu}}
\maketitle              
\vspace{-4mm}
\begin{abstract}
Federated learning (FL) is a recently developed area of machine learning, in which the private data of a large number of distributed clients is used to develop a global model under the coordination of a central server without explicitly exposing the data. The standard FL strategy has a number of significant bottlenecks including large communication requirements and high impact on the clients' resources. Several strategies have been described in the literature trying to address these issues.  
In this paper, a novel scheme based on the notion of ``model growing" is proposed. Initially, the server deploys a small model of low complexity, which is trained to capture the data complexity during the initial set of rounds. When the performance of such a model saturates, the server switches to a larger model with the help of {\it function-preserving transformations}. 
The model complexity increases as more data is processed by the clients, and the overall process continues until the desired performance is achieved. Therefore, the most complex model is broadcast only at the final stage in our approach resulting in substantial reduction in communication cost and client computational requirements. The proposed approach is tested extensively on three standard benchmarks and is shown to achieve substantial reduction in communication and client computation while achieving comparable accuracy when compared to the current most effective strategies.

\end{abstract}

\vspace{-7mm}
\section{Introduction}
Federated learning (FL) is a new machine learning (ML) paradigm that enables the training of a model by utilizing private data distributed across many clients governed by a central server \cite{li2020federated}. In contrast to the traditional ML, where all samples are stored in a single place, FL assumes that the data are generated and collected by many distributed, independent clients. Therefore, the overall data is expected to be heterogeneous and non-IID (Identically and Independently Distributed).  
In training, the server broadcasts the current global model to a set of randomly selected clients. Each selected client locally trains the received model with its private data and sends the updates to the server. The server aggregates the updates on the current model using federated averaging. This constitutes a single communication round. This procedure is repeated for many communication rounds, where in each round the randomly selected clients advance the training, until convergence is achieved. FL has been growing in importance especially with the emergence of edge computing and AI on the edge due to the massive deployment of IoT devices and advances in communication and networking systems \cite{yu2021toward, savazzi2020federated}.

In the most general setting, the standard FL strategy has a number of significant bottlenecks that need to be addressed before it can be widely used in practice. These bottlenecks include data heterogeneity \cite{diao2020heterofl}, unreliable and variable rate connectivity \cite{li2020federated}, uneven and relatively limited client resources \cite{imteaj2021survey}, high communication requirement, and high impact on the clients' resources. Strategies to address these bottlenecks is suggested in the literature, see for example \cite{kairouz2019advances} but here we focus our attention on the two bottlenecks of communication requirements and the limited client resources. 

Three approaches have been suggested to reduce the communication requirement and achieve comparable performance to the standard FL. 
The first, as in \cite{mao2021communication, xu2020ternary, elkordy2022heterosag}, relies on quantization methods; the second relies on the sparsification of the model update as in \cite{konevcny2016federated}; and the third approach broadcasts smaller networks to improve communication efficiency \cite{li2021lotteryfl,wu2021fedkd}.

Another constraint is the relatively limited resources available at the clients. In general, whenever a state-of-the-art model is required to capture the global heterogeneous data complexity, FL induces a significant computational overhead on the clients. 
Therefore, we should aim at reducing the computational requirements on the clients. In \cite{caldas2018expanding}, federated dropout is introduced in which random sub-networks of the entire model are broadcast thereby reducing communication bandwidth and computational resources. 

The main contributions of this paper are:
\begin{itemize}
    \item A novel strategy called FedNet2Net is introduced in which we start with a small initial model, and gradually enlarge the model to capture the increasing complexity of the data processed by the clients and improve accuracy. 
    \item Function preserving transformations are used to switch from one model to the next once the performance of the current model saturates. Our switching is efficient and ensures a continuous improvement in accuracy as long as the inherent complexity of the data increases.
    \item  FedNet2Net is shown through extensive experiments to result in large savings in the amounts of computation and communication compared to several of the best known strategies.
    \item FedNet2Net can be used to adaptively terminate at the smallest possible model that achieves the desired accuracy.
\end{itemize}

The rest of the paper is organized as follows. Section 2 provides an overview of techniques related to reducing the communication and client resources, while Section 3 describes our approach in details. Section 4 introduces the benchmarks used for evaluation and describes the details of our model implementations. We present and discuss the results in Section 5 and we conclude in Section 6.

\vspace{-4mm}
\section{Related Work}
Since the introduction of federated averaging \cite{mcmahan2016communication}, reducing communication bandwidth and local computation has received a great deal of interest in the community. 
Most of the work tried to improve communication efficiency, which can be broadly categorized into model compression based techniques \cite{mao2021communication, xu2020ternary, elkordy2022heterosag ,zheng2021distributed, li2021communication, qiao2021communication, pmlr-v119-rothchild20a}, update sparsification based techniques \cite{konevcny2016federated} or broadcasting smaller networks \cite{li2021lotteryfl,jiang2019model,wu2021fedkd}. Model compression techniques involve mainly quantization \cite{mao2021communication, xu2020ternary, elkordy2022heterosag}, compressed sensing \cite{li2021communication}, low rank approximation \cite{qiao2021communication} and tensor decomposition \cite{zheng2021distributed}. The approach in 
\cite{konevcny2016federated} combines sparsification of gradient updates with quantization for reducing the client-to-server communication cost. Model pruning technique has been used for communicating smaller networks, for example by applying the lottery ticket hypothesis for pruning \cite{li2021lotteryfl, jiang2019model}. 
The work in \cite{liang2020think} partitions the model into local parameters for representation learning and global parameters for broadcasting by training the whole model locally. 
Significant savings is not expected from models having fully connected (FC) layers at the bottom as they possess most of the parameters of the entire model.
Moreover, for testing new client data, the scheme needs to pass the data through all local models and ensemble the outputs, which is not suitable in practice. 
Some of the works avoid communicating the entire model by replacing it with either logits from model outputs as in \cite{ itahara2020distillation, sattler2020communication} or binary masks as in \cite{li2021fedmask}.

The above schemes mostly improve communication efficiency but do not deal directly with possible computation savings at the client side. To achieve this objective, the typical approach 
used is mainly focused on broadcasting a subnetwork to clients. For example, 
federated dropout \cite{caldas2018expanding} and adaptive federated dropout \cite{adaptive_federated_dropout} randomly drop out a fixed percentage of units or filters to broadcast a sub-network combined with lossy compression at each round. 
Unlike the other knowledge distillation based methods, the approach in \cite{he2020group} trains small models on clients and transfers their knowledge to a large server-side model. The approach used in 
\cite{li2021fedmask} only communicates and learns personalized binary masks, while freezing the model parameters. This idea, however, restricts the model to update its weights. On the other hand, the method described in 
HeteroFL \cite{diao2020heterofl} broadcasts submodels to clients based on their computation capability assuming that this knowledge is present to the server. 
All the discussed methods consider a single global model throughout the entire training. 
We just found out that a concurrently developed method in \cite{progfed2021} progressively adds randomly initialized layers (or blocks) from the final model architecture with new classification layers at specific intervals. We expect this scheme to result in a drop of performance at the beginning of each stage and hence, we believe its performance will be significantly lower than ours.

\section{Proposed Approach}
 
We develop a modified training scheme to make the local training computationally much less demanding and reduce the communication requirement while maintaining accuracy. 
We observe that to achieve a state-of-the-art accuracy, a model needs to capture the essential characteristics of the global data held in the clients, which are known only after the participation of a large number of clients. However, it is not necessary to broadcast such a model throughout the entire training. At the beginning, the model starts with a small amount of data available at relatively few clients, and then progresses with additional information after each round of FL. 
Intuitively, a model with much lower complexity can be deployed at the beginning and then, we can exploit transfer learning strategies to transfer the knowledge from a lower complexity model to a higher complexity model. In this way, the complexity of the model grows as the data complexity increases, until we reach a final model that can capture the global data.

Our approach is based on transfer learning, but in a different way than standard transfer learning in which the top layers of a student network are directly copied from a teacher network \cite{Yosinski2014how}.  
Instead, the strategy of function-preserving transformations introduced in \cite{chen2015net2net} is utilized, where the student network is initialized in such a way that it represents the same function as the teacher but with different parameterization. We start federated training with a small network. As the training proceeds and more data from clients are exposed, the model is enlarged using two transformations. The first is Net2Widernet, which replaces a model with an equivalent model that is wider, i.e. the student model will have a larger number of units at a certain layer(s). The second operation is Net2DeeperNet, which replaces a model with an equivalent deeper model. These two transformations constitute the essential components of Net2Net. Each transformation preserves the function of the network. After initializing the student network that has the same function as the teacher network, the student network is trained further to improve performance, and once no improvement is detected, the network is enlarged using the Net2Net transformations. We call our scheme FedNet2Net after the Net2Net methodology. Details are presented next.

\begin{figure}[!t]
    \centering
    \includegraphics[width=0.90\linewidth]{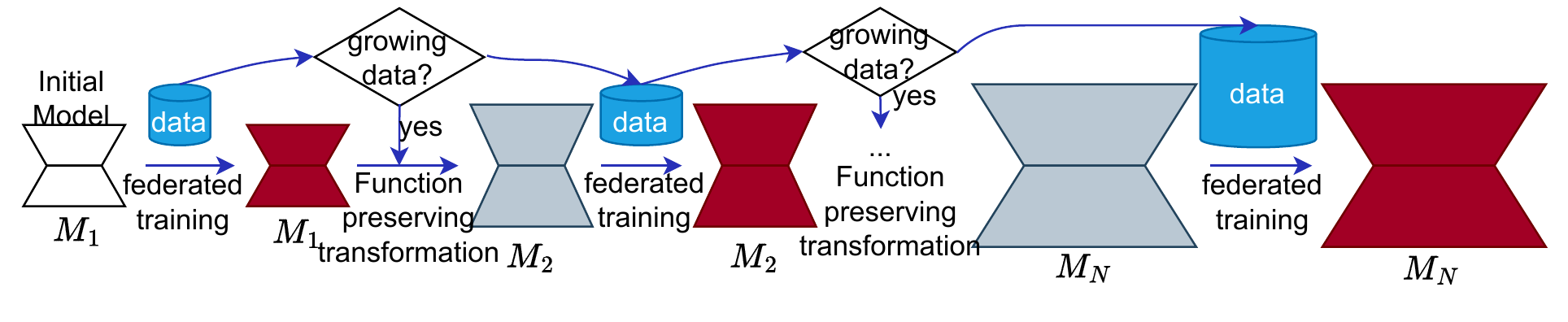}
    \caption{Overview of the FedNet2Net training.}
    \label{fig:overview}
\end{figure}

\begin{figure}[!t]
    \centering
    \includegraphics[width=0.40\linewidth]{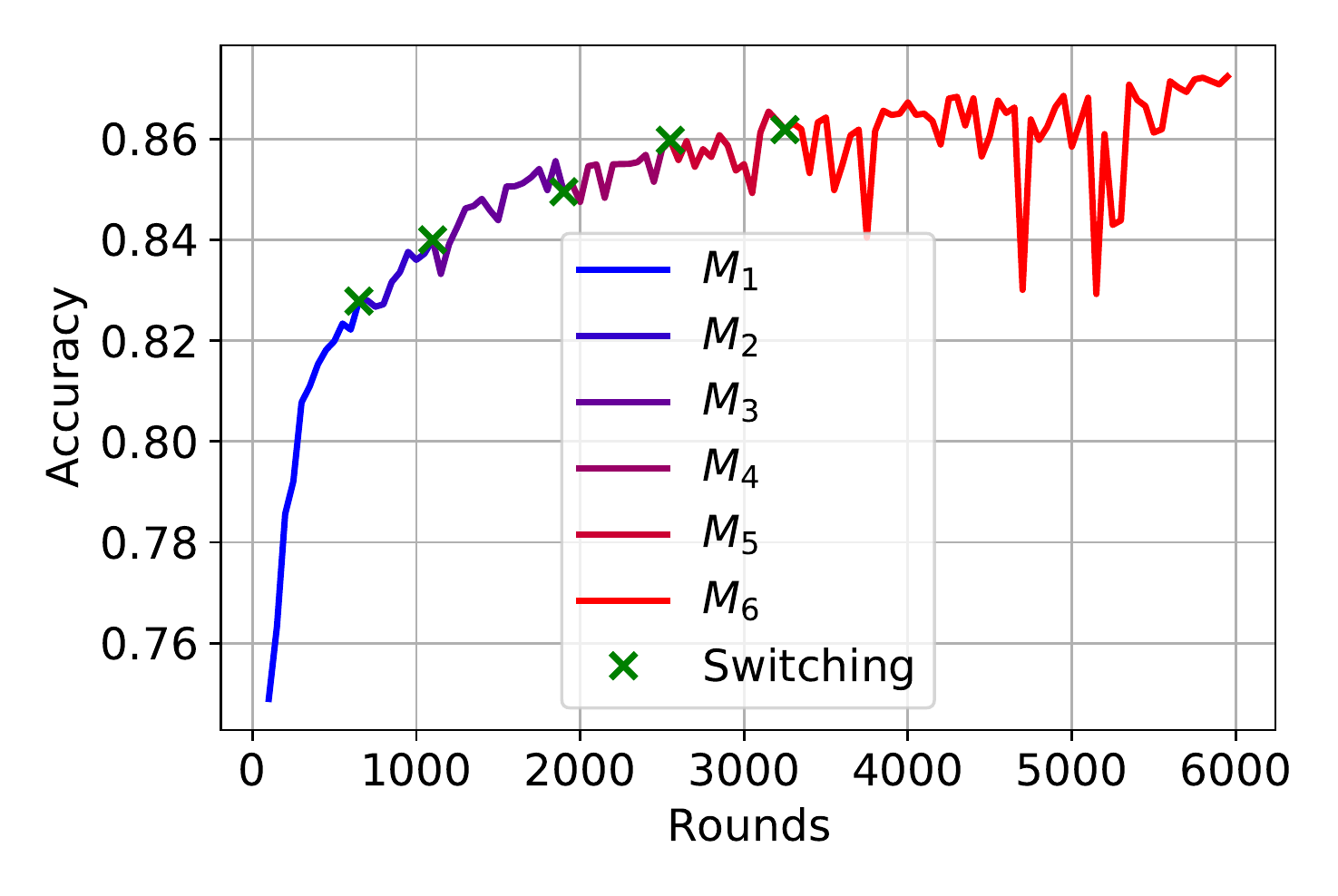}
    \caption{Model switching in FedNet2Net training using the EMNIST dataset}
    \label{fig:FedNet2NetTrain}
\end{figure}

Let us assume model $1$ to model $N$, $\{M_1, M_2, ..., M_N\}$ are of increasing complexity, where $M_1$ is a model having the fewest number of layers and units. $M_N$ is a model that can capture the entire data fairly well, and could have the same architecture as deployed in the standard federated training. $M_1$ through $M_N$ are designed in such a way that for two consecutive models $M_i$ and $M_{i+1}$, the architecture of $M_{i+1}$ has either increased the number of layers or increased the number of units in some layer(s) or both compared to $M_i$, as enlarging the model width or depth is proven to be effective in improving accuracy \cite{he2020group}. Fig. \ref{fig:overview} presents the overview of FedNet2Net training.
During each stage, the model is deployed after enlarging it by applying a functional preserving transformation to meet the growing data complexity without degradation in performance. In this way, we continue to train a model until its progress saturates after which we switch to the next model. Fig. \ref{fig:FedNet2NetTrain} shows the model switching in FedNet2Net training.
We observe that since the highest complexity model is broadcast only at the final stage, we achieve a large reduction in communication bandwidth and clients' computational requirements  while achieving similar performance as in the standard FL. 
In a federated setting, FedNet2Net can be a life long learner and can adapt over a long period of time \cite{silver2013lifelong}.

We now describe how the transformation is function preserving and how to apply the transformation to transfer knowledge from $M_i$ to $M_{i+1}$. The student network $M_{i+1}$ has either increased the number of layers or increased the number of units in some layer(s) compared to $M_i$. In the case of Net2DeeperNet, to insert a FC layer to $M_i$, we initialize the weights of the inserted layer as the identity matrix. To insert a convolutional layer, we initialize the kernel as the identity filter. 
Similarly, Net2WiderNet operation replaces a layer with a wider layer while preserving the function value, that is more units for FC layers, or more filters for convolutional layers. 
For a convolutional kernel $K_l$ of shape $(w_l,h_l,i_l,o_l)$, $w_l$ and $h_l$ denote filter width and height, and $i_l, o_l$ are the number of input and output channels of layer $l$. To widen this kernel $\hat{K}_l$ having a shape of $(w_l,h_l,i_l,\hat{o}_l)$, where $\hat{o}_l> o_l$, a random mapping is considered as follows:
\vspace{-2mm}
$$G_l(j)=\begin{cases}
    j,& \text{if } 1 \le j \le o_l\\
    \text{random sampling from } \{1,2,...,o_l\},  & \text{if } o_l \le j \le \hat{o}_l
\end{cases}$$

Then the new kernel becomes $\hat{K}_l[x,y,i,j]= K_l[x,y,i,G_l(j)]$. Therefore, the first $o_l$ number of output channels are directly copied while the rest of the output channels are randomly sampled and copied to the new kernel.
For preserving the function value, the next layer kernel $K_{l+1}$ is also reduced due to the replication in its
input. The new kernel $\hat{K}_{l+1}$ is given as $\hat{K}_{l+1}[x,y,j,k]=\frac{K_{l+1}[x,y,G_l(j),k]}{|\{z|G_l(z)=G_l(j)\}|}$. Similar transformation can be realized for FC layers. More details regarding the transformations appear in \cite{chen2015net2net}. 
This operation preserves the function value, that is, for any sample, the function value of the model before and after the transformation remains the same. As we have dropout layers in our models, no extra noise adding is needed for advancing the training since the dropout technique achieves the same goal.

We now address the issue of when to switch to a larger model.

\subsubsection{Switching Policy.}
For the server to decide when to switch from one model to the next, we adopt a switching policy based on the model loss. As the loss at each round fluctuates significantly due to training and random selection of clients, we calculate the window-based loss at $t$-th round and then compare it to another window of a certain time lag $L$ as follows. 
\vspace{-0.2cm}
\begin{equation}
S_t=\frac{1}{N}\sum_{i=1}^{N}\text{loss}[t-i-L]-\frac{1}{N}\sum_{i=1}^{N}\text{loss}[t-i] \label{eqn_loss}
\end{equation}
where $\text{loss}[i]$ is the weighted average of local model losses from clients at $i$-th round, $L$ is the time lag between two consecutive windows of size $N$. In other words, we compute the running average of losses for $N$ rounds of the earlier past and the recent past and take the difference. 
This measure captures the progress in training based on the loss over the time window. If $S_t$ is not above a certain threshold, then switching to a larger model is performed. Given that the clients are selected randomly during each round, our measure to use the average loss over a window stabilizes the switching decision process. A new model is trained for at least $N+L$ rounds before determining the next switching decision. 
The thresholds are tuned to get the best possible results.
A similar switching policy can be realized  by using the validation accuracy evaluated at the server; however this assumes that there is a public validation dataset available at the server.

\section{Datasets and Detailed Model Implementations}
In this section, we present brief descriptions of the datasets and the detailed model sequences used to conduct various experiments. 

\subsection{Data Description}
In order to evaluate the proposed training approach, three benchmark datasets namely, EMNIST, CIFAR-10, and MNIST are used. For federated training, the training sets in MNIST and CIFAR10 are randomly divided into 100 clients. 
The EMNIST dataset is an extension of the MNIST, which consists of 671585 images of 62 classes split into 3400 unbalanced non-IID clients \cite{caldas2018leaf}.

\subsection{Performance Evaluation}
To evaluate FedNet2Net, we implement and compare the following five methods.
\begin{enumerate}
\item \textbf{\textit{FedAvg:}} Federated averaging \cite{mcmahan2016communication} with traditional dropout layer. Here, we add a dropout layer \cite{srivastava2014dropout} after each convolutional and FC layer. Dropping out of units or filters is performed inside clients. 
\item \textbf{\textit{FD:}} Federated dropout \cite{caldas2018expanding}. Because of broadcasting smaller subnetworks, some amount of communication and computation are saved compared to FedAvg. We consider FedAvg and FD as the baseline methods.
\item \textbf{\textit{HeteroFL}} \cite{diao2020heterofl}: We uniformly sample computation level of each client at each round from 5 different levels.  
\item \textbf{\textit{FedNet2Net (FNN):}} This is our approach with traditional dropout layer.
\item \textbf{\textit{FedNet2Net-FD (FNN-FD):}} Our approach is combined with federated dropout to reduce further communication and computation. Here, dropout is applied to all models except the smaller ones.
\end{enumerate}

For performance evaluation, we plot the accuracy against total communication of all methods and the reduction in average communication per round versus the accuracy of our approaches over the baselines. The second plot can be used to determine the communication saved for a desired accuracy, and to determine the number of clients that can participate at a round. While not mentioned explicitly, similar reduction in the amount of computations is achieved since the models used are much smaller for the majority of the rounds (if not all).
All methods are trained for 6000 rounds and for convenience, the test accuracy is recorded after every 50 rounds.

\begin{table}[b]
  \centering
  \caption{Thresholds for five consecutive switchings and hyperparameters for three datasets.}
    \begin{tabular}{c|c|c|c}
    \hline
dataset & Thresholds for policy (\ref{eqn_loss}) & learning rate & clients per round \\\hline\hline
EMNIST &  $[0.08, 0.04, 0.02, 0.01, 0.005]$ & 0.035 & 35\\ \hline
CIFAR10 &  $[0.12, 0.11, 0.10, 0.09, 0.08]$ & 0.05 & 10\\ \hline
MNIST & $[0.04, 0.02, 0.01, 0.005, 0.0025]$ & 0.015 & 10\\ \hline

    \end{tabular}%
  \label{threshold_switching}%
\end{table}%

\vspace{-2mm}
\subsection{Parameters for Switching}
As mentioned before, we switch from one model to a larger one when we detect that the training improvement is below a certain threshold over a running window, as described by the equation (\ref{eqn_loss}). We use $N=100$ and $L=300$ for all datasets. The thresholds for consecutive switching are listed in Table \ref{threshold_switching}. The switching is decided at the server, and hence, no extra computation or communication is incurred at the client side.

\begin{table}[tbh]
  \centering
  \caption{Consecutive models of FedNet2Net training for the EMNIST and MNIST datasets. The number of units are written in parenthesis. For MNIST, we use the same architectures except each classification layer has 10 units and the number of units in the second last layer is 128 for models 1-4, 256 for model 5 and 512 for model 6. Kernel size is $5 \times 5$.}
    \begin{tabular}{p{1.5cm}||p{9.1cm}|p{1.2 cm}}
    \hline
Model number & Architecture & Para- meters \\\hline\hline
model 1 & Conv2D(16) $\rightarrow$ maxpool(4,4) $\rightarrow$ FC(512) $\rightarrow$ FC(62) & 434K\\ \hline
model 2 & \textbf{Conv2D(32)} $\rightarrow$ maxpool(4,4) $\rightarrow$ FC(512)  $\rightarrow$ FC(62) & 836K\\ \hline
model 3 & Conv2D(32) $\rightarrow$ maxpool(2,2) $\rightarrow$ \textbf{Conv2D(32)} $\rightarrow$ maxpool(2,2) $\rightarrow$ FC(512) $\rightarrow$ FC(62) & 862K\\ \hline
model 4 & Conv2D(32) $\rightarrow$ maxpool(2,2) $\rightarrow$ \textbf{Conv2D(64)} $\rightarrow$ maxpool(2,2) $\rightarrow$ FC(512) $\rightarrow$ FC(62)  & 1.7M\\ \hline
model 5 & Conv2D(32) $\rightarrow$ maxpool(2,2) $\rightarrow$ Conv2D(64) $\rightarrow$ maxpool(2,2) $\rightarrow$ \textbf{FC(1024)} $\rightarrow$ FC(62) & 3.3M\\ \hline
model 6 & Conv2D(32) $\rightarrow$ maxpool(2,2) $\rightarrow$ Conv2D(64) $\rightarrow$ maxpool(2,2) $\rightarrow$ \textbf{FC(2048)} $\rightarrow$ FC(62) & 6.6M\\
\hline
    \end{tabular}%
  \label{modelsMnist}%
\end{table}%

\begin{table}[htb]
  \centering
  \caption{Consecutive models in FedNet2Net for CIFAR10. Kernel size is $3 \times 3$ everywhere except the last two convolutional layers of model 6, where it is $1 \times 1$.}
    \begin{tabular}{p{1.5cm}|p{9.2cm}|p{1.2 cm}}
    \hline
Model number & Architecture  &  Para- meters \\\hline\hline
model 1 & Conv2D(32) $\rightarrow$ maxpool(3,3) $\rightarrow$ Conv2D(64) $\rightarrow$ maxpool(3,3) $\rightarrow$ Conv2D(10) $\rightarrow$ GlobalAveragePooling2D (GAP) $\rightarrow$ FC(10) & 20K\\ \hline
model 2 & Conv2D(32) $\rightarrow$ \textbf{Conv2D(32)} $\rightarrow$ maxpool(3,3) $\rightarrow$ Conv2D(64) $\rightarrow$ \textbf{Conv2D(64)} $\rightarrow$ maxpool(3,3) $\rightarrow$ Conv2D(10) $\rightarrow$ GAP $\rightarrow$ FC(10) & 66K\\ \hline
model 3 & \textbf{Conv2D(64)} $\rightarrow$ \textbf{Conv2D(64)} $\rightarrow$ maxpool(3,3) $\rightarrow$ \textbf{Conv2D(128)} $\rightarrow$ \textbf{Conv2D(128)} $\rightarrow$ maxpool(3,3) $\rightarrow$ Conv2D(10) $\rightarrow$ GAP $\rightarrow$ FC(10) & 262K\\ \hline
model 4 & \textbf{Conv2D(96)} $\rightarrow$ \textbf{Conv2D(96)} $\rightarrow$ maxpool(3,3) $\rightarrow$ \textbf{Conv2D(192)} $\rightarrow$ \textbf{Conv2D(192)} $\rightarrow$ maxpool(3,3) $\rightarrow$ Conv2D(10) $\rightarrow$ GAP $\rightarrow$ FC(10) & 586K\\ \hline
model 5 & Conv2D(96) $\rightarrow$ Conv2D(96) $\rightarrow$ maxpool(3,3) $\rightarrow$ Conv2D(192) $\rightarrow$ Conv2D(192) $\rightarrow$ maxpool(3,3)  $\rightarrow$ \textbf{Conv2D(192)} $\rightarrow$ Conv2D(10) $\rightarrow$ GAP $\rightarrow$ FC(10) & 918K\\ \hline
model 6 & Conv2D(96) $\rightarrow$ Conv2D(96) $\rightarrow$ maxpool(3,3) $\rightarrow$ Conv2D(192) $\rightarrow$ Conv2D(192) $\rightarrow$ maxpool(3,3)  $\rightarrow$ Conv2D(192) $\rightarrow$ \textbf{Conv2D(192)} $\rightarrow$  Conv2D(10) $\rightarrow$ GAP $\rightarrow$ FC(10) & 955K\\
\hline
    \end{tabular}%
  \label{modelsCifar}%
\end{table}%

\vspace{-2mm}
\subsection{Model Description and Hyper-parameters}
The hyperparameters and models for FedAvg, FD and the final stage of FedNet2Net are set as suggested in \cite{caldas2018expanding}. The learning rates and the number of clients per round are listed in Table \ref{threshold_switching}. The number of epochs per round is set to 1 and the batch size for local training is 10. SGD optimizer and sparse categorical cross-entropy loss are used. Dropout rate is set to 0.125.
The sequence of models used in FedNet2Net are presented in Tables \ref{modelsMnist} and \ref{modelsCifar} along with the number of parameters in each model. Changes in consecutive models are highlighted. RELU activation and dropout is applied after each convolutional and FC layer except the last classification layer, where a softmax activation is applied.

\begin{figure*}[!t]
    \centering
    (a) EMNIST dataset\\
    \includegraphics[width=0.42\linewidth]{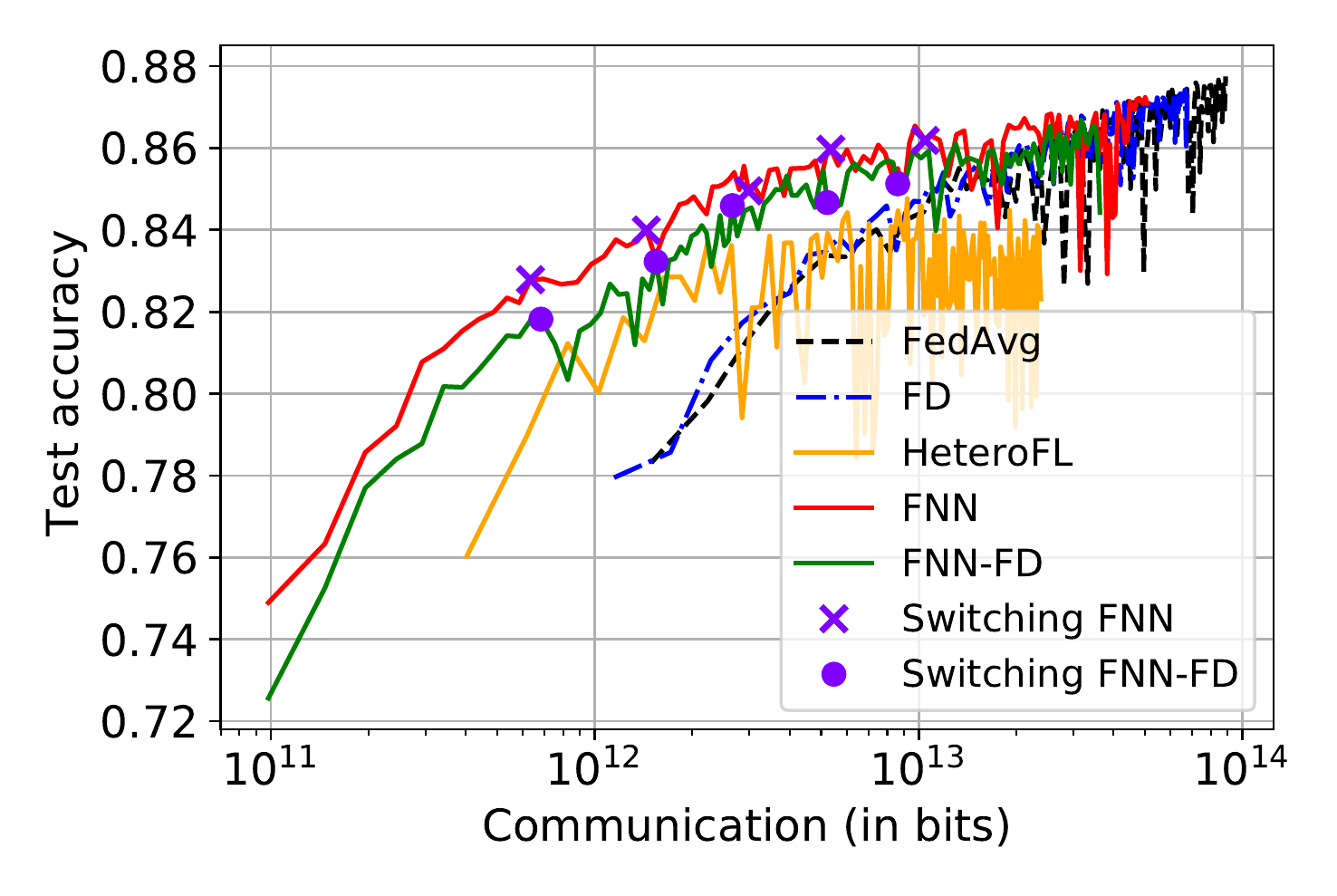}
    \includegraphics[width=0.41\linewidth]{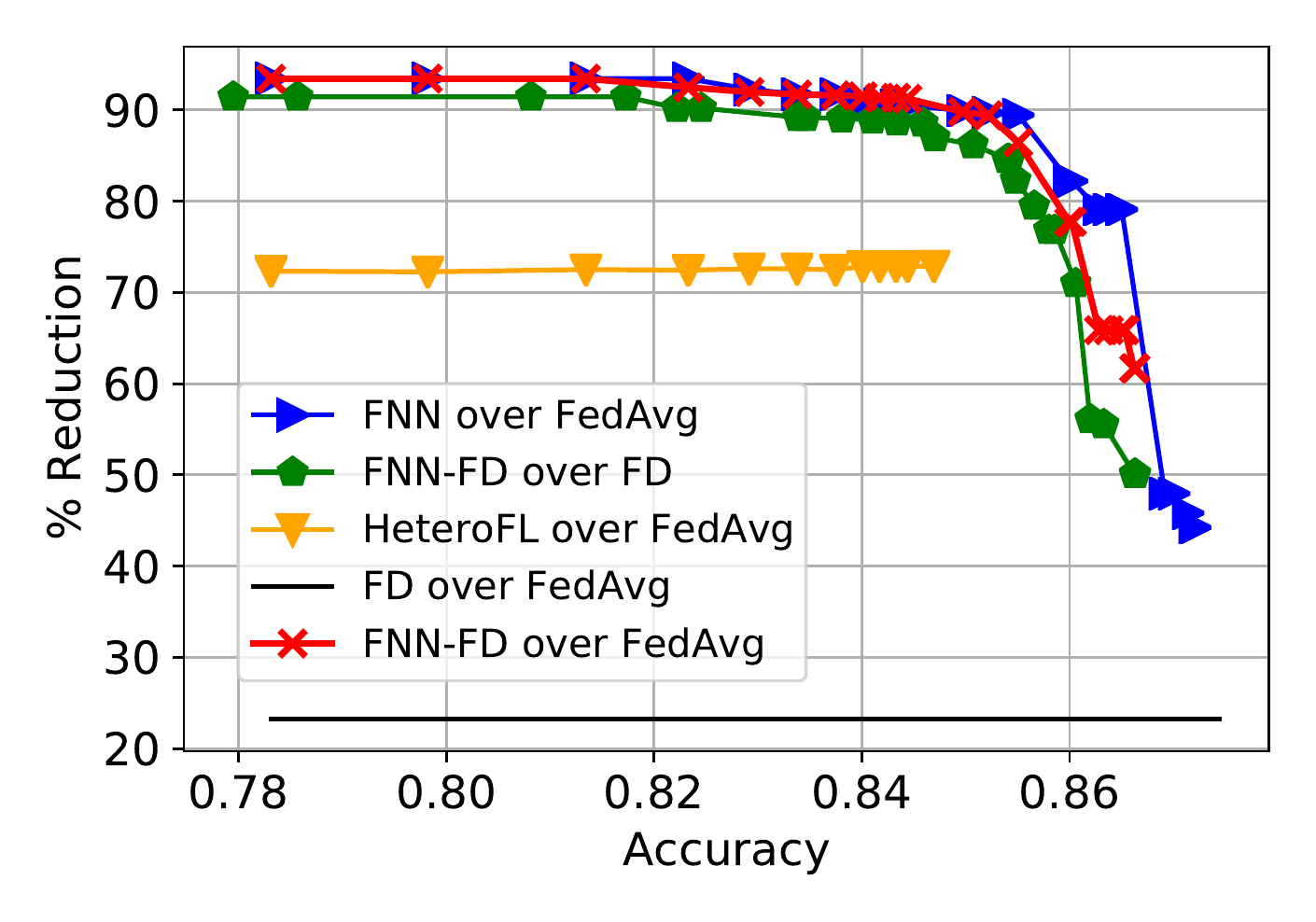}\\
    (b) CIFAR10 dataset\\
    \includegraphics[width=0.41\linewidth]{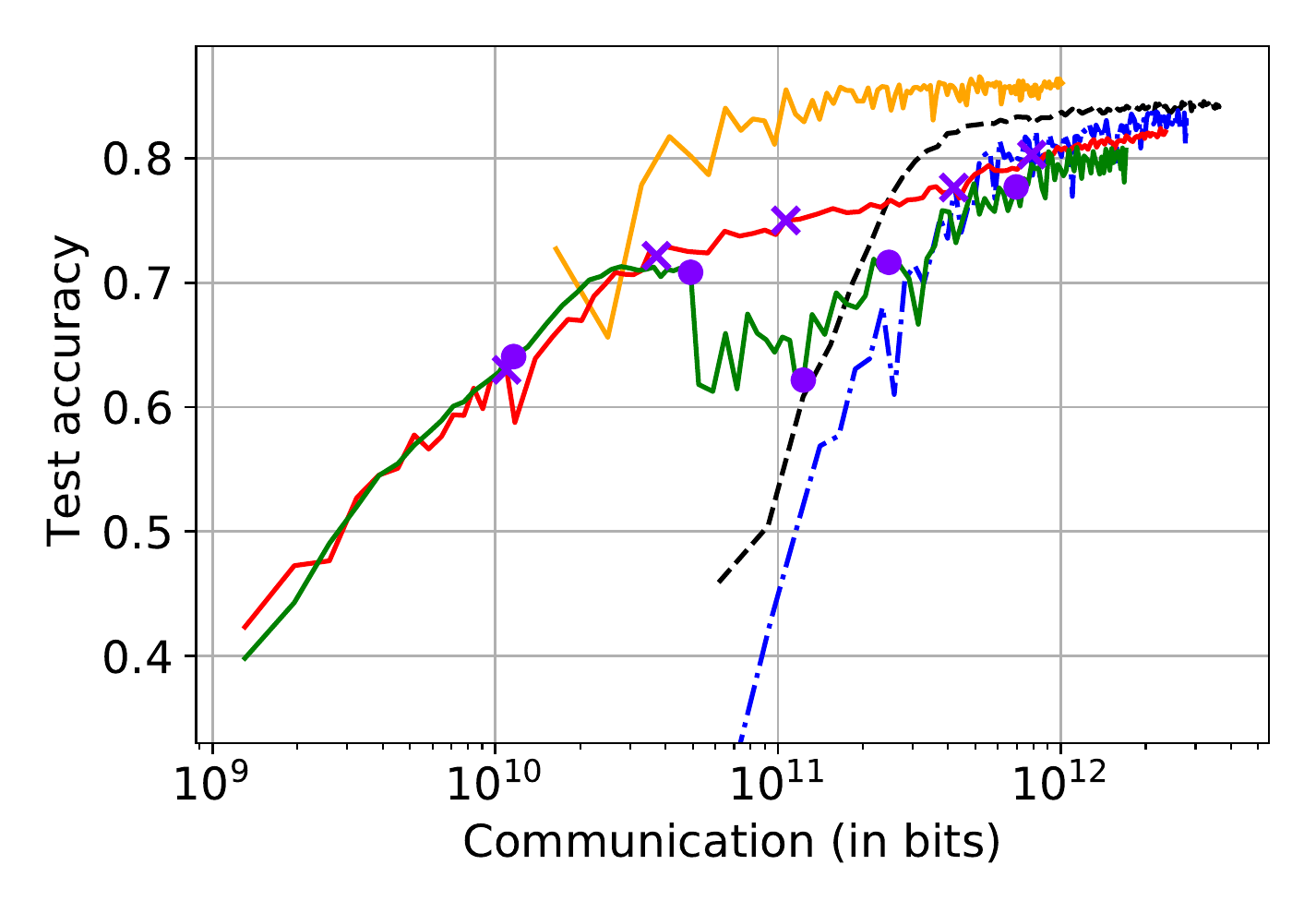}
    \includegraphics[width=0.41\linewidth]{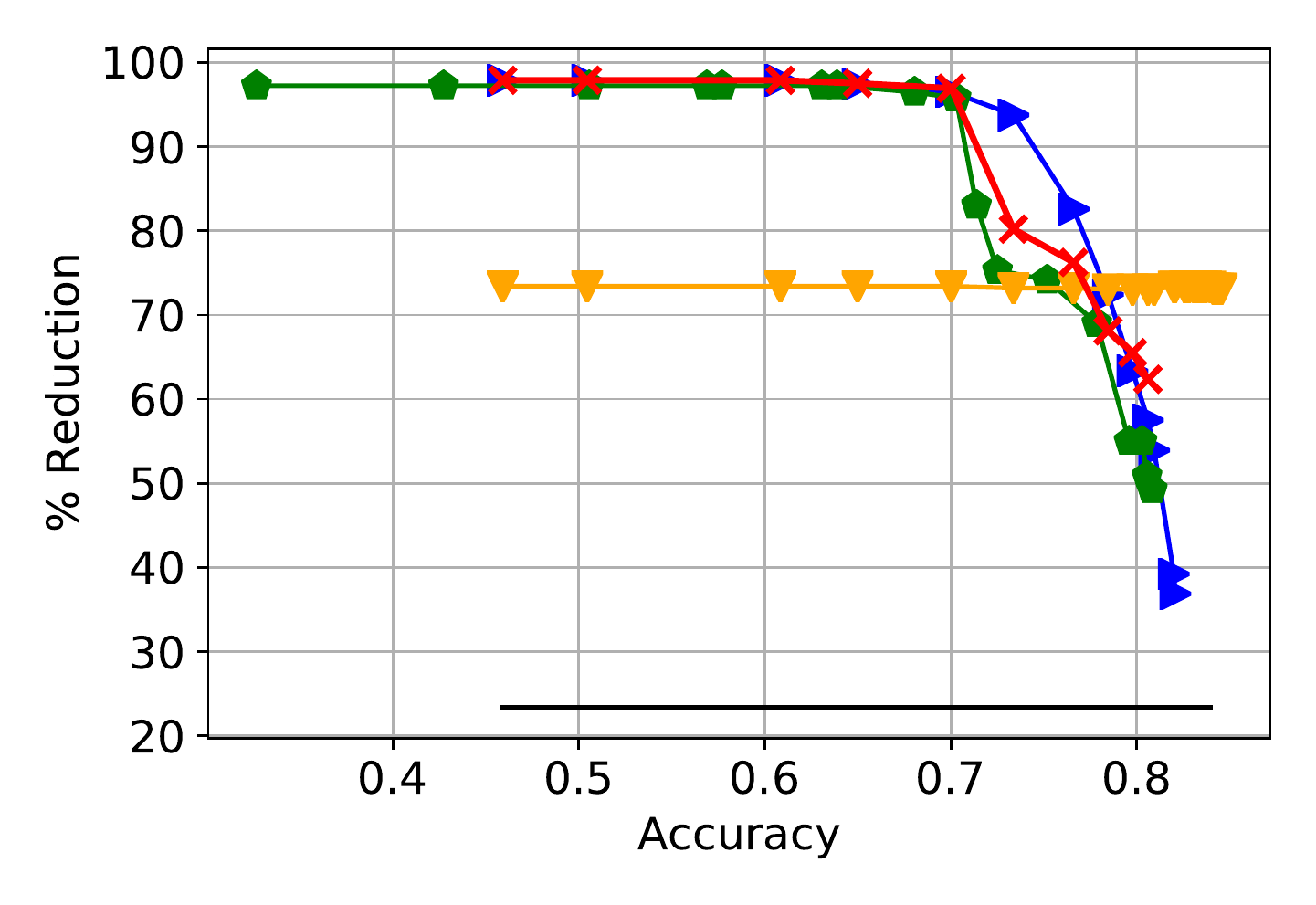}\\
    (c) MNIST dataset\\
    \includegraphics[width=0.42\linewidth]{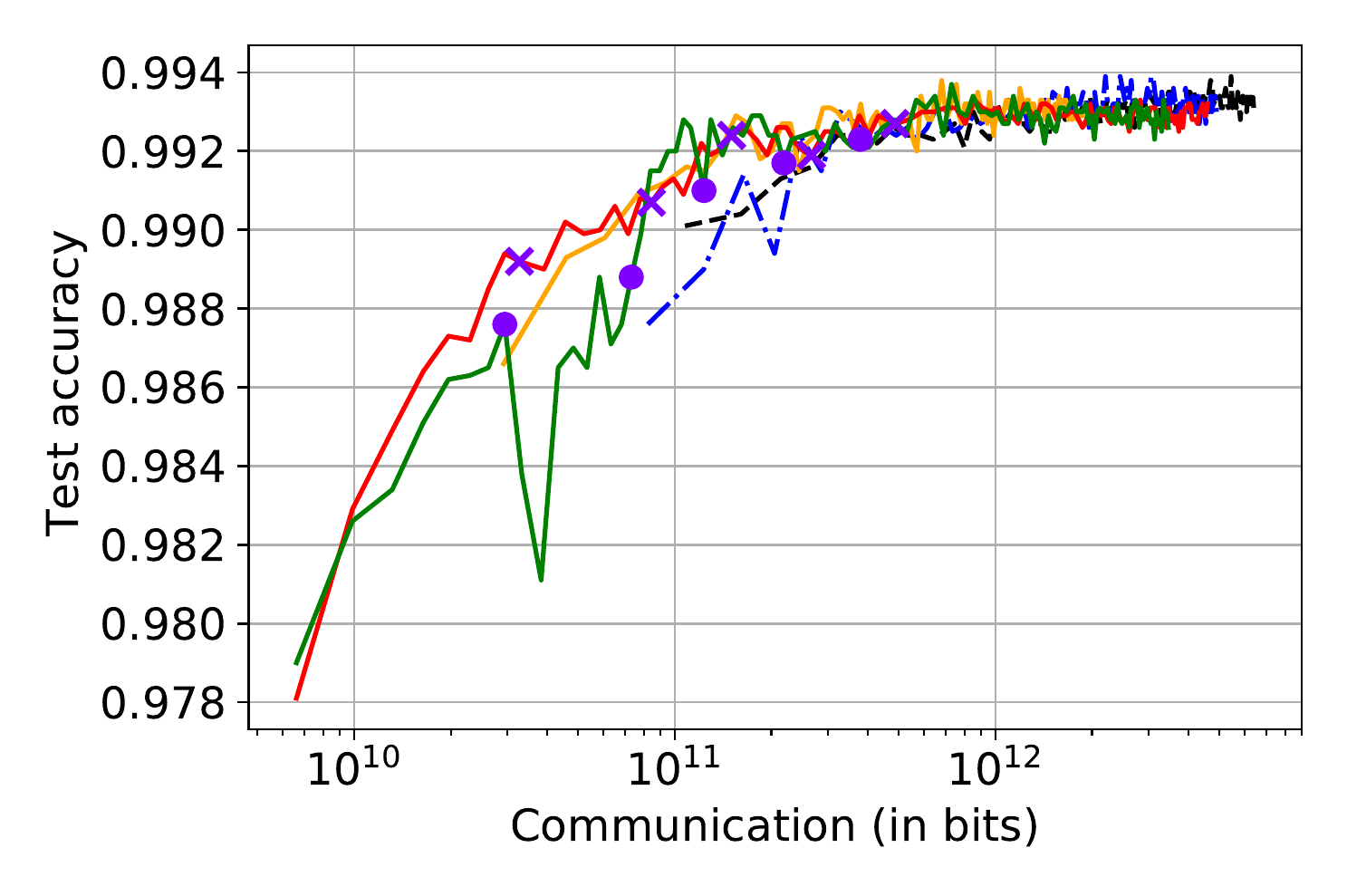}
    \includegraphics[width=0.41\linewidth]{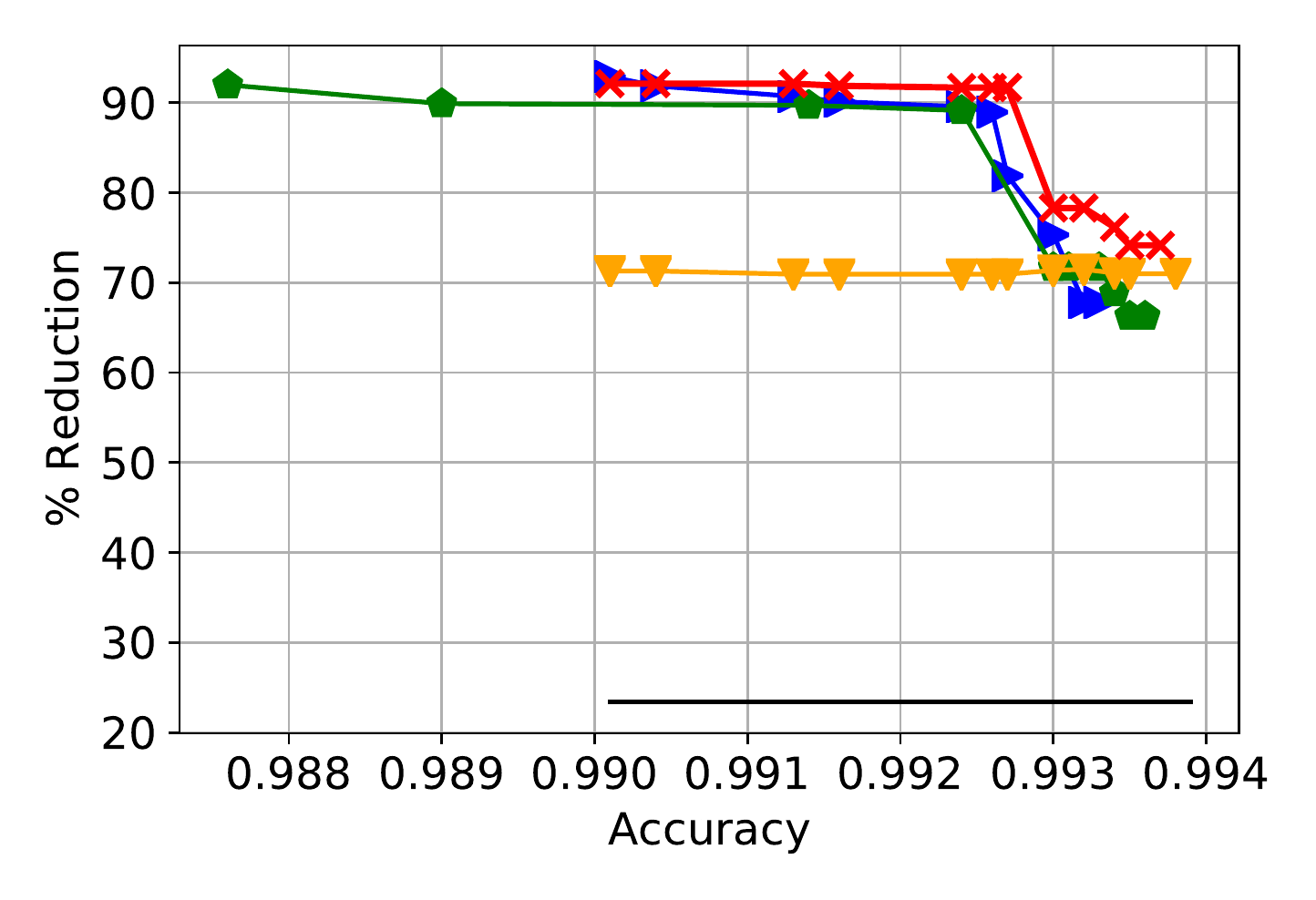}
    \caption{Accuracy against communication of the FedAvg, FD, HeteroFL, FedNet2Net and FedNet2Net-FD (left), and percentage reduction in communication per round of FedNet2Net, FD and HeteroFL over FedAvg (right). The markings in the left figures indicate the switching positions based on policy (\ref{eqn_loss}). Accuracies fluctuate due to training from randomly selected clients at each round. }
    
    \label{fig:loss}
\end{figure*}

\section{Results}
Fig. \ref{fig:loss} compares the performance between FedNet2Net and the baseline methods using the model loss based switching policy. 
The figures on the left present the accuracy against total communication (in bits) with switching positions in our approach. We observe that for any fixed amount of communication, FedNet2Net achieves higher and occasionally equal accuracy compared to FedAvg, federated dropout and HeteroFL; the only exception is the performance of HeteroFL on CIFAR10 at the high accuracy end, which we believe is due to adding batch normalization layers in HeteroFL, which boosted the training. Otherwise, HeteroFL training is highly unstable.

The figures on the right present the percentage reduction in communication per round of FNN over FedAvg, FNN-FD over FD, FD over FedAvg, HeteroFL over FedAvg and FNN-FD over FedAvg.
It is observed that, at the slightly lower accuracy regions than the best possible, the percentage savings is approximately more than 90\% per round. This saving remains constant for most of the accuracy regions. At the higher accuracy region, the percentage savings starts to drop, which intuitively makes sense as we are switching to larger models to capture the growing data complexity. Moreover, the reduction of FD and HeteroFL over FedAvg remains the same for all accuracies as they broadcast roughly the same number of parameters at all rounds. However, additional per round communication savings is achieved by combining FedNet2Net with FD. 
Similar reduction is achieved when we separated a validation set from the training data to measure validation accuracy for making switching decisions.
Therefore, our training approach based on model growing reduces significant amount of communication per round. Moreover, as our approach starts from a significantly small model and gets transformed into a higher capacity model, it opens the opportunity to adapt the model as data complexity grows. 

\vspace{-0.2cm}
\subsubsection{Effect of choosing the different set of intermediate models.}
\begin{figure*}[!t]
    \centering
    \vspace{-0.2cm}
    \includegraphics[width=0.41\linewidth]{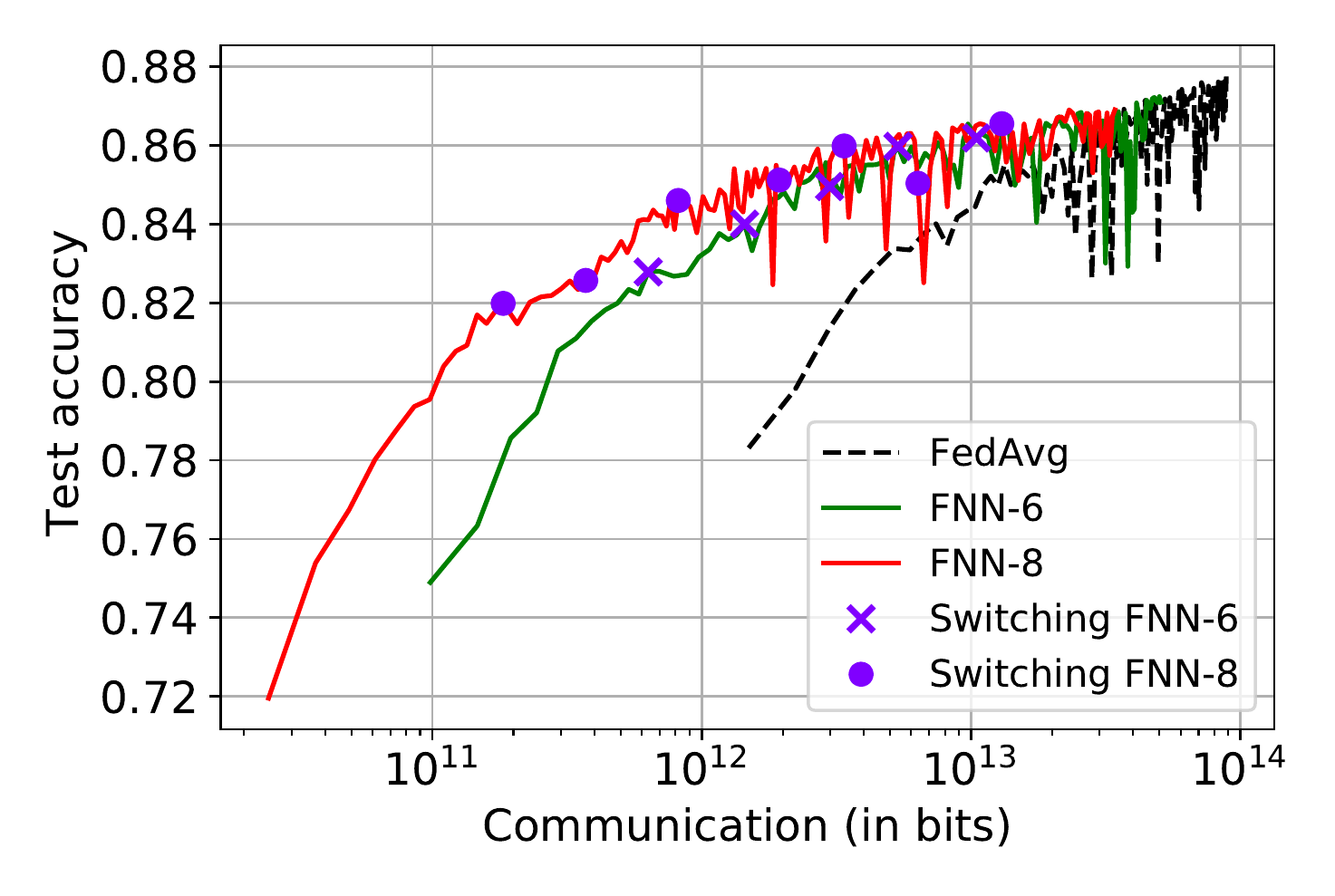}
    \includegraphics[width=0.41\linewidth]{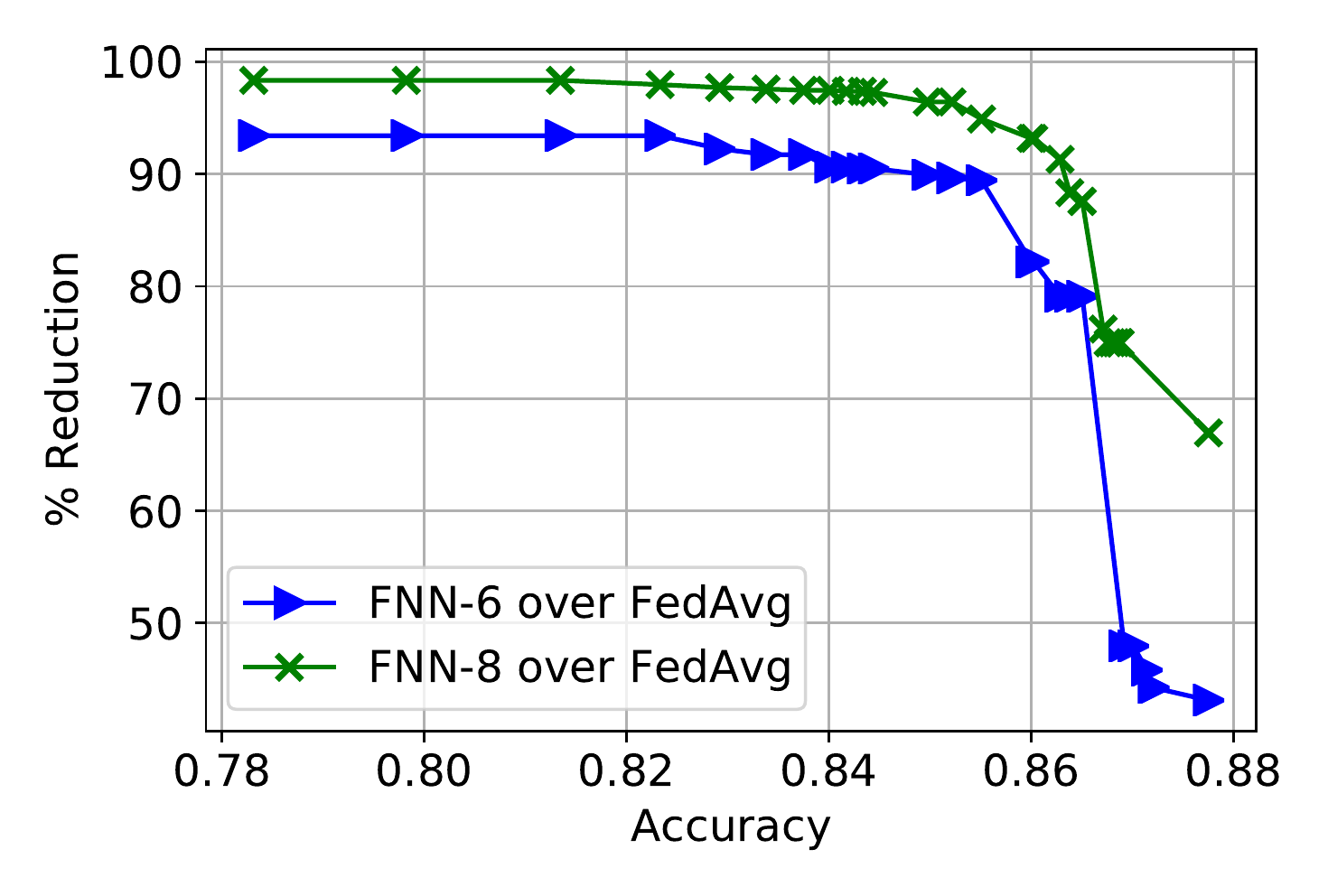}
    \caption{Performance comparison of FedNet2Net with different set of intermediate models for the EMNIST dataset.}
    \label{fig:diff_steps}
\end{figure*}

Given the final model, we implement FedNet2net with a different set of intermediate models to show the effect of choosing them differently. For EMNIST, we implement FedNet2Net with sets of 6 and 8 models (FNN-6 and FNN-8), and the results are presented in Fig. \ref{fig:diff_steps}. FNN-8 has a smaller starting network than FNN-6. We observe that FNN-8 performs better than FNN-6 in terms of communication reductions. This is intuitive as the final model is broadcast for lesser number of rounds. Similarly for CIFAR10, we implement FedNet2Net by changing some of the intermediate models mentioned in Table \ref{modelsCifar}. The reductions achieved is almost the same as in Fig. \ref{fig:loss}(b). We omitted the figure due to space limitations.

\vspace{-0.2cm}
\section{Conclusion}
In this paper, a new federated training scheme, based on the model growing strategy, is proposed for saving both communication cost and computations at the clients. At the initial stage, as the model is trained using only a small amount of data, deploying a model with minimal capacity saves both communication and local computation. 
Next, the efficient switchings to the enlarged models using function-preserving transformations ensure a continuous improvement in performance as long as the complexity of the data increases.
In this way, the capacity of the model is increased until the model captures the overall data complexity. 
The most complex model is deployed only at the final stages of training, and thereby, saving a substantial amount of communication and local computations. 
Extensive experiments on three benchmarks demonstrate that the proposed FedNet2Net training scheme can save significant amount of communication cost and local computations per round.

%
%
\bibliographystyle{splncs04}
\bibliography{FedNet2Net_CR}
\end{document}